\documentclass[11pt]{article}

\usepackage[preprint]{acl}

\usepackage{times}
\usepackage{latexsym}

\usepackage[T1]{fontenc}

\usepackage[utf8]{inputenc}

\usepackage{microtype}

\usepackage{inconsolata}

\usepackage{graphicx}

\usepackage{dirtytalk}
\usepackage{amsmath}
\usepackage{enumitem}

\title{Introducing the Privacy-HSD Trade-off:\\Hate Speech Detection, but not at the Cost of Privacy}

\author{
 \textbf{Stephen Meisenbacher\textsuperscript{1,2,3}},
 \textbf{Vlad Garbuz\textsuperscript{4}},
 \textbf{Chirill Donos\textsuperscript{5}},
 \textbf{Maxim Dnestreanschii\textsuperscript{4}},
\\
 \textbf{Gabriel Creanga\textsuperscript{4}},
 \textbf{Andreea-Elena Bodea\textsuperscript{1}},
 \textbf{Thomas Lampert\textsuperscript{6}},
 \textbf{Jana Diesner\textsuperscript{1,2,3}}
\\
 \textsuperscript{1}Technical University of Munich, Chair of Human-Centered Computing, Munich, Germany\\
 \textsuperscript{2}Munich Center for Machine Learning, Munich, Germany\\
 \textsuperscript{3}Munich Data Science Institute, Munich, Germany\\
 \textsuperscript{4}University of Southern Denmark, Sønderborg, Denmark\\
 \textsuperscript{5}Technical University of Moldova, Chișinău, Republic of Moldova\\
 \textsuperscript{6}ICube, University of Strasbourg, Strasbourg, France
\\
 \small{
   \textbf{Correspondence:} \href{mailto:stephen.meisenbacher@tum.de}{stephen.meisenbacher@tum.de}
 }
}

\begin{document}
\maketitle
\begin{abstract}
Hate speech is a real and timely threat that affects a large portion of online users, especially youth and minority groups. While building reliable and robust automatic hate speech detection (HSD) systems is paramount, we argue that this must also be balanced with the individual right to privacy. Exploring the intersection of HSD and privacy, we demonstrate that HSD systems might unintentionally achieve performance at the cost of encoding authorship, posing a threat to privacy. Building on these findings, we establish the notion of a \textit{privacy-HSD trade-off}, which demands a careful balance. We benchmark a series of text privatization methods, as well as our newly proposed domain-specific \textsc{AgnoSpeech} technique, showing that balancing privacy and HSD is difficult but feasible. The findings make a strong case for more research on the trade-offs between privacy and HSD, both of which have tangible implications for the safeguarding of online participation.
\end{abstract}

\section{Introduction}
As the venue of social discourse continues to shift from the public sphere to online environments, the threat of online hate speech has unfortunately also grown rapidly in prevalence \cite{alkomah2022literature,gandhi2024hate}. A recent 2025 UNESCO report\footnote{\scriptsize\url{http://unesco.org/en/world-media-trends}} found that nearly 67\% of online users have encountered hate speech at some point, and in the European Union, this number is nearly 50\% among young adults (aged 16 to 29)\footnote{\scriptsize\url{https://ec.europa.eu/eurostat/web/products-eurostat-news/w/ddn-20240801-1}}. Alarmingly, the United Nations has found that almost 70\% of individuals who personally experience online hate are part of a minority group\footnote{\scriptsize\url{https://www.un.org/en/hate-speech/impact-and-prevention/targets-of-hate}}. These statistics paint a reality in which hate speech has become a serious and pervasive online threat, for which immediate mitigations are required.

In response to the increasing threat of online hate speech, parliamentary bodies such as the Council of Europe have released recommendations for combating hate speech, such as in the recent CM/REC(2022)16\footnote{\scriptsize\url{https://search.coe.int/cm?i=0900001680a67955}}, as an integral part of protecting the rights and freedoms of individuals. In the academic literature, both predating and following this recommendation, the response has been immense. Automatic approaches to hate speech detection (HSD), powered by Natural Language Processing and more recently LLMs, have grown in research attention \cite{10.1145/3232676,alkomah2022literature,gandhi2024hate,10848067}. The diversity in which such approaches have addressed hate speech has been the subject of several recent surveys (ibid), and they often center on HSD in the context of social media platforms \cite{10025718,rawat2024hate}.

A novel and unexplored angle in the pursuit of automatic HSD comes with its implications on \textit{privacy}. In other words, automatic and reliable HSD should ideally not depend or infringe upon the right to privacy of individuals. The CM/REC(2022)16 hints at this dilemma, calling for \say{removing such hate speech without delay}, but doing so while \say{respecting privacy and data-protection requirements}. This connection between HSD and privacy is intuitive yet significant and complex to balance, in the sense that accurate hate speech detectors should not achieve their reliability by detecting identifying authorship cues rather than hate speech signals. Such a relationship grounds itself in similar, well-studied trade-offs that must be balanced with privacy preservation, most notably the \textit{privacy-utility trade-off} \cite{li2012theories,slavkovic2023statistical}. In this way, the \textit{privacy-HSD} trade-off is conceived. 

We position our work in the foundation and initial exploration of the privacy-HSD trade-off. Firstly, we perform a probing study with two representative datasets, highlighting the degree to which HSD classification systems may also unintentionally encode authorship cues. Secondly, we formalize the privacy-HSD trade-off and conduct a case study to investigate current text privatization techniques in improving this trade-off. Finally, we design a novel, HSD-specific privatization method (\textsc{AgnoSpeech}), which we benchmark against methods from the literature for effectiveness in managing the privacy-HSD trade-off.

Our findings reveal that HSD performance is inherently entangled with privacy considerations, particularly in encoding author-specific information. In addressing this privacy concern, our tailored \textsc{AgnoSpeech} privatization method achieves higher trade-offs than off-the-shelf text privatization methods, the latter of which are unable to reliably balance HSD with privacy, while also producing coherent and usable outputs. These results make the case for domain- and use case-specific text privatization, especially in critical domains such as HSD where privacy implications matter.
In addition to the above, we make the following contributions:

\begin{enumerate}
    \itemsep 0em
    \item We address and formalize the privacy implications of HSD, leading to the establishment and study of the privacy-HSD trade-off.
    \item We demonstrate the entanglement of HSD and privacy, showing that unmitigated HSD methods may inadvertently serve as profiling tools.
    \item We propose a novel text privatization method, \textsc{AgnoSpeech}, that preserves hate speech cues for use in building HSD tools, while also selectively removing signals not necessary for HSD. The replication code and data can be found at: \url{https://github.com/AllForOne-md/agnospeech-core}
    \item We evaluate \textsc{AgnoSpeech} and comparative methods on the privacy-HSD trade-off with three datasets for privacy-preserving HSD.
\end{enumerate}

\section{Related Work}
\paragraph{Hate Speech Detection.}
The topic of automatic Hate Speech Detection has produced a rich and diverse field of research over the past decade, coinciding with the rising threat posed by growing hate speech found on the internet. These numerous approaches have been reviewed and systematized by multiple recent surveys \cite{schmidt-wiegand-2017-survey,10.1145/3583067,gandhi2024hate,rawat2024hate,malik2025deep}.

While we do not conduct an additional survey, we acknowledge that the breadth of HSD literature is too vast to enumerate here, and we therefore discuss a representative sample that follows the progression of the technical work of the past decade. Earlier approaches leveraged static resources, such as lexicons, to identify texts containing harmful or hateful language \cite{gitari2015lexicon,davidson2017automated}.
Ensuing works leveraged the growing promise of Machine Learning and Deep Learning techniques to build accurate detectors of hate speech \cite{10.1145/2740908.2742760,10.1145/2872427.2883062,park-fung-2017-one,10.1145/3357384.3358040,9139953}. With the rise of transformer-based language models, more recent works have explored their usefulness in HSD, which often achieve superior performance in detection \cite{mozafari2019bert,toraman-etal-2022-large,10459901,yang-etal-2023-hare,ghorbanpour-etal-2025-prompting}.

A large focus has also been placed on producing datasets and benchmarks for HSD research \cite{rottger-etal-2021-hatecheck,mathew2021hatexplain,hartvigsen-etal-2022-toxigen}, as well as identifying important considerations therein \cite{10.1145/3331184.3331262}. In addition, other streams of research have branched HSD to multilingual capabilities \cite{rottger-etal-2022-multilingual,arango-monnar-etal-2022-resources,chhabra2023literature,10662891}, as well as to multimodal settings \cite{9093414,hee-etal-2024-recent,bui-etal-2025-multi3hate}. Recent surveys also highlight the metrics utilized to evaluate automatic HSD, which predominantly include precision-recall-F1 setups or AUC measures, either in a binary (hate/non-hate) or more nuanced multi-class setting \cite{9455353,chhabra2023literature,gandhi2024hate}.

\paragraph{Text Privatization.}
The field of \textit{text privatization} is likewise varied and diverse, but it is united in the goal of redacting, masking, or otherwise transforming textual data in a manner to remove direct or indirectly identifiable information \cite{ren-etal-2025-measure}. In the literature, this has taken the form of many classes of techniques, including anonymization, text scrubbing, and private text rewriting \cite{11247969}. As previously mentioned, one of the primary challenges of text privatization is the reliable removal of personal identifiers, while also maintaining the utility and integrity of the data \cite{meisenbacher-etal-2024-comparative}.

The exact execution of text privatization can vary depending on the underlying privacy notion. Classic anonymization typically targets personally-identifiable information (PII) and related named entities \cite{11247969}, where these are either removed or masked. More recent works in anonymization have leveraged LLMs for this task \cite{bao-carpuat-2024-keep,yang-etal-2025-robust,frikha-etal-2025-incognitext}. Other works operate on more subtle clues hidden within text, such as indirect identifiers \cite{baroud-etal-2025-beyond} like medical conditions or socioeconomic status, or writing style and stylometry \cite{10.1145/2382448.2382450}. Further sub-fields focus on text privatization under formal privacy notions, such as Differential Privacy, which obfuscate and transform text with mathematical privacy guarantees \cite{klymenko-etal-2022-differential,carvalho2023tem,igamberdiev-habernal-2023-dp}. While all of the works typically address the impact of privatization on the privacy-utility trade-off, none have done so by interpreting the utility aspect as that of HSD. 

The evaluation of text privatization approaches is an important, yet complex consideration, which necessitates thinking adversarially in order to measure mitigation effectiveness against such adversaries. Prior work has noted the difficulties of defining \say{privacy} in textual data \cite{10.1145/3531146.3534642,ren-etal-2025-measure}. Nevertheless, recent efforts have been made to systematize and standardize such evaluations \cite{pilan-etal-2022-text,loiseau-etal-2025-tau,huang-etal-2025-nap2}. 
One salient use case for text privatization, which is discussed further below is \textit{protection against re-identification}, which we study here, namely to quantify privacy preservation in protecting anonymous online users from being identified via their data.

\paragraph{Authorship Obfuscation.}
Akin to the field of text privatization, \textit{authorship obfuscation} focuses specifically on removing author-identifying signals from written texts, in order to mitigate the risk of downstream \textit{authorship re-identification}. Classic approaches from both obfuscation and identification engineer numerous stylistic features from text, which can be used to profile (or mask) authorship \cite{potthast2016author}. While such feature-based approaches usually rely on heuristics for obfuscation \cite{bevendorff-etal-2019-heuristic}, more modern techniques have turned to neural representations and LLM-based obfuscation \cite{fisher-etal-2024-styleremix,10.1145/3715073.3715076,shokri-etal-2025-personalized}. \citet{loiseau-etal-2025-tarot} frames the task of authorship obfuscation in light of the resulting trade-off with utility; in our work, we extend this notion to the task of HSD.

While re-identification is typically seen as adversarial, the tasks of \textit{authorship attribution} (e.g., identify a historical text's author) \cite{stamatatos-2017-authorship,tan-etal-2025-open} and \textit{authorship verification} (verify if two texts were written by the same person) \cite{hung-etal-2023-wrote,bevendorff-etal-2025-two} can be positive and useful, and profiling may even assist in detecting online abuse \cite{mishra-etal-2018-author}. Our work, though, studies the negative aspects of authorship attribution: placing online users at risk of re-identification when building HSD systems.

\section{Dataset Construction}
To aid in the experiments and analyses of this work, we curate two base datasets from online forums, which importantly contain both user (author) and hate speech labels. We leverage previously published research datasets and filter these down to constrained author sets for use in our experiments.

\paragraph{Reddit.}
We use a dataset provided by \citet{qian-etal-2019-benchmark}, which covers 22k Reddit comments from 5k conversations, which have been annotated as hate or non-hate speech, based on a keyword analysis. As noted by the original work, we reiterate that the dataset does not necessarily contain all of the comments attached to a post, but rather those preceding and trailing an identified hate speech comment. Only those comments matching the keyword search were marked as hate speech; all others receive a negative (0) label.

We retain the comments associated with the top-25 most frequently posting users in the dataset, resulting in 1154 comments, 365 (31.6\%) of which are marked as hate speech. The most frequently contributing user appears 100 times, while the least frequent user among the top 25 has 32 comments.

We also create a subset with the top-50 authors to support our analysis in Section \ref{sec:foundations}. It contains 1795 comments, 525 (29.2\%) of which are hate speech, and a least-contributing user with 21 comments.

\paragraph{Twitter.}
We also used a dataset of 16k tweets provided by \citet{waseem-hovy-2016-hateful}, where each tweet is marked as racist, sexist, or neither. We simplify this to hate and non-hate speech, to match the annotation format of the Reddit subset. Using the X API (\url{https://docs.x.com/}), we successfully re-hydrated 9,523 of the 16k tweets with the original text content and associated user IDs, where the non-successful attempts imply that the original posts have been deleted or moved.

We narrow down the subset to the top-10 most frequent authors, which comprises 6,792 tweets. Among these, 2039 (30\%) are tagged as hate speech. The subset is heavily skewed towards the top-3 (3626, 2040, and 929 tweets, respectively), with the 10th user only contributing 25 tweets.

\paragraph{License and release.}
All datasets will be publicly released upon publication. The Reddit datasets will be distributed as created, enabled by the original data source's license. However, due to the terms of the Twitter dataset, specifically in re-hydrating the tweets, we only will release a version without the user IDs and tweet texts. Researchers who wish to obtain these must do so through the Twitter (X) API, using the tweet IDs as the reference point.

\section{Foundations of the \texorpdfstring{$PrivHSD$} \ \ Trade-off}
\label{sec:foundations}
Using our three created datasets, we conduct an initial investigation into the potential for HSD to encode authorship information and thus partially devolve into a re-identification tool. From these results, we motivate the need for \textit{privacy-preserving HSD}, in which detection performance does not come at the cost of degrading privacy protections.

\subsection{Training HSD Models}
To support our study of the privacy-HSD trade-off, we train three baseline HSD binary classification models, two for Reddit (25 and 50 user subsets) and one for Twitter. To train a more robust classifier, the entire original cleaned datasets (22k for Reddit and 9.5k for Twitter) are used; however, a 20\% random sample is held-out from the top-k author subsets. These held-out sets are treated as the test sets for model performance and the ensuing analysis.

For all setups, we fine-tune a \textsc{google-bert/bert-base-cased} model \cite{devlin-etal-2019-bert}, chosen due to its popularity \cite{tucudean2024natural}, using the Hugging Face Trainer library. Reddit models are trained for one epoch and Twitter for three epochs (due to the smaller training size). A learning rate of 1e-5 and Adam optimizer are used. All training is performed with a max model input length of 128 (i.e., otherwise truncated) and batch size of 64 on a single Nvidia RTX 5060 Ti 16GB GPU.
On the held-out set, the Reddit-25 model achieves a micro-F1 score of 0.89, the Reddit-50 model a score of 0.86, and the Twitter model a score of 0.91. All three models are  available at \url{https://hf.co/collections/sjmeis/privhsd}.

\begin{table*}[t]
\centering
\resizebox{\linewidth}{!}{
\begin{tabular}{c|ccc|ccc|cc|}
 & \multicolumn{3}{c|}{\textbf{Linear Probing}} & \multicolumn{3}{c|}{\textbf{Statistical Tests}}  & \multicolumn{2}{c|}{\textbf{Adversary}}\\ \cline{2-9} 
\multicolumn{1}{c|}{Dataset / HSD Model} & \multicolumn{1}{c}{Random/Majority} & \multicolumn{1}{c}{Baseline} & \multicolumn{1}{c|}{Probe Acc. (HSD)} & \multicolumn{1}{c}{$\eta^2$} & \multicolumn{1}{c}{Mean FPR} & \multicolumn{1}{c|}{FPR $\sigma$} & \multicolumn{1}{c}{Majority Class} & \multicolumn{1}{c|}{Micro F1}\\ \hline
 Reddit-25 & 4\%/8.2\% & 40.69\% & 39.83\% & 0.2205 & 7.68\% & 0.1079 & 15.95\% & 19.04\% \\
 Reddit-50 & 2\%/4.5\% & 25.35\% & 21.73\% & 0.2227 & 8.33\% & 0.1564 & 10.55\% & 13.09\% \\
 Twitter-10 & 10\%/53.1
 \% & 83.96\% & 88.24\% & 0.6095 & 16.36\% & 0.2078 & 69.61\%  & 87.20\%
\end{tabular}
}
\caption{Results of our linear probing and statistical tests. Probe accuracy measures the ability to train authorship re-identification models solely from HSD model embeddings, as compared to random or majority-guessing accuracy (on the test set), as well as the non-fine-tuned baseline. $\eta^2$ denotes variance in HSD that can be attributed to the author signal, where as FPR $\sigma$ indicates the standard deviation of false positive rates across all authors in a dataset. The Adversary (F1) denotes adversarial classification performance (for authorship), trained solely on text contents. In all scores, a higher result denotes that authorship has become entangled with detecting hate speech.}
\label{tab:probe}
\end{table*}

\subsection{Probing the HSD Models}
Using the fine-tuned HSD models, we perform two analyses to identify whether the learned hate speech signal (i.e., for binary classification) has become intertwined with authorship-revealing information.

\paragraph{Linear probing.}
Using the trained HSD models, we probe their internal representations to test whether the trained weights have encoded author-identifying information. This is done per model:

\begin{enumerate}[leftmargin=1.5em]
    \itemsep 0em
    \item For each text input from the train set, extract the \textsc{[CLS]} token embedding from the last hidden state, i.e., that was used by the classification head for binary hate speech classification.
    \item With these 768-dim. embeddings, train a multinomial Logistic Regression model (\textsc{sklearn}, \texttt{max\_iter}=1000) to predict the author (label).
    \item Obtain the embeddings from the held-out test set texts, and measure the accuracy of the regression model in predicting the author.
\end{enumerate}

We also followed this procedure on the base, non-fine-tuned \textsc{BERT}. In performing these steps, we are able to measure to what degree the trained HSD model's internal representations are correlated with author identity, i.e., the authors from the test set in question. The results of this probing study are provided in Table \ref{tab:probe}, which show that for all three dataset configurations, authorship is inherently intertwined with the model's internal representations, where the trained regressor achieves far greater than random or majority chance accuracies in predicting authorship. The HSD model's learned representations either persist (Reddit) or exacerbate (Twitter) this issue, showing that training HSD models without privacy in mind may leave online users at risk.

\paragraph{Statistical measures.}
In addition to linear probing of the HSD models, we also perform statistical tests to measure the correlation of HSD and author.

Firstly, we measure the correlation between each HSD model's prediction confidence (i.e., of the binary hate speech label) to the categorical author label. This is calculated using a one-way ANOVA test to achieve the $\eta^2$ score. To calculate this, we extract the prediction probabilities (post-softmax, interpreted as a confidence score) from the held-out test set inputs. Then, we train an Ordinary Least Squares regression model (using \textsc{statsmodel}). Finally, we calculate $\eta^2$, or the proportion of the variance in the regression model's predictions that is attributable to the author ID. As shown in Table \ref{tab:probe}, for all models, the calculated effect sizes are all considerably high, particularly for Twitter.

Probing more deeply into \textit{individual} impacts, we measure per-author error rate to measure the entanglement of model error and authorship. For this, we focus on the false positive rates (FPR) in the model predictions on the test set. Reported first in Table \ref{tab:probe} are the mean FPR across all authors in each dataset. We then report the standard deviations of these FPRs, showing the distribution spread across authors. Higher standard deviations imply that a HSD model is more prone to commit false positives for certain authors, and thereby, the trained models may be more tuned to authorship style than to hate speech signals. A truly fair and privacy-preserving model would have deviations close to 0, but as evidenced in Table \ref{tab:probe}, this is not the case.

\subsection{Modeling a Capable Adversary}
Rather than rely only on post-hoc analyses using the trained HSD models, we conclude our probes with the modeling of a capable adversary who directly trains a re-identification model to classify authorship given an input text. This represents a black-box, more plausible scenario in which an adversary has access to the publicly known texts of a set of users (but not the trained HSD models), and trains a classification model to re-identify future texts, potentially where the author label is hidden.

Similar to HSD, we train \textsc{google-bert/bert-base-cased} models on the train splits of our prepared datasets, but instead of predicting the hate speech label, the objective is now to classify the author label, i.e., in a constrained multi-class setting. The re-identification performance is represented by the trained model's micro-F1 score on the test set. We use the same training setup as before, with the exception of 15 epochs for Reddit and five epochs for Twitter (due to the smaller training sets).

The results are also shown in Table \ref{tab:probe}. We see that re-identification scores represent modest, but non-negligible improvements over majority class guessing, especially considering the black-box nature of the attack. These results, combined with those above, lend evidence to the idea that HSD performance and privacy preservation are intertwined.

\subsection{The Privacy-HSD Trade-off}
To formalize a trade-off between HSD performance and privacy protection, we adopt the notion of \textit{relative gain} from the privacy literature \cite{mattern-etal-2022-limits}, which weighs relative losses in utility (here, HSD performance) with gains in privacy protection (here, defense against re-identification). We measure this balance over the \textit{baseline}, which are the HSD and privacy scores on non-privatized datasets.

Given the performance of a trained HSD classifier on the original datasets ($H_o$) and the performance of the same classifier configuration after being trained on the privatized dataset counterpart ($H_p$), we define the relative utility change to be the following, which is normalized against HSD performance (F1) with majority-class guessing:
\begin{displaymath}
    \Delta HSD = \frac{H_p - H_{maj}}{H_o - H_{maj}}.
\end{displaymath}

Similarly, for a set of privacy metrics $\mathcal{P}$, all normalized between 0 and 1, with 0 being \say{ideal} privacy, we define the relative privacy change as:
\begin{displaymath}
    \Delta Privacy = \frac{1}{n} \sum \frac{m_p}{m_o}, \forall m \in \mathcal{P}, n = |\mathcal{P}|.
\end{displaymath}

In our work, we use four metrics $m \in \mathcal{P}$: probe accuracy, $\eta^2$, FPR $\sigma$, and adversarial F1 (see Table \ref{tab:probe}), which all denote better privacy when minimized.
Following this, we define the privacy-HSD trade-off as the relative gain between the two quantities:
\begin{equation}
    PrivHSD =  \Delta HSD - \Delta Privacy
    \label{eq:to}
\end{equation}

This trade-off typically lies in the range [-1, 1], although it can exceed these bounds if HSD performance is better post-privatization, or if privacy is worse than the original data. A higher $PrivHSD$ score implies that privacy gains outweigh utility losses (left side is dominant), and vice versa.

\section{A Case Study of Existing Mitigations}
Given the newly defined $PrivHSD$ trade-off metric, we conduct a case study on a selection of recent text-to-text privatization methods, benchmarking their ability to balance the trade-off. Each selected method, introduced below, is run on all three curated datasets, in order to produce a privatized counterpart (the \say{text} is replaced with a private version). These private dataset counterparts are used to fine-tune corresponding HSD models, which are in turn evaluated using the $PrivHSD$ criteria.

\subsection{Selected Privatization Methods}
We select seven state-of-the-art text privatization methods, which range from three overarching research areas: entity-based anonymization, differentially private text privatization, and LLM-assisted anonymization. For all but one method, we run them on two configurations, which allows for a representative sweep of their privatization capabilities.

\paragraph{Microsoft Presidio.}
The Presidio tool\footnote{\scriptsize\url{https://github.com/data-privacy-stack/presidio}} has two components, an \textit{analyzer} and an \textit{anonymizer}, which identify and handle detected personally identifiable information, respectively. We use it in two configurations: (1) \textit{redact}, where detected entities are deleted, and (2) \textit{replace}, where the entities are replaced with a placeholder, e.g., \say{[PERSON]}.

\paragraph{\textsc{GLiNER}.}
\textsc{GLiNER} \cite{zaratiana-etal-2024-gliner} is a framework for deploying Named Recognition (NER) models, which are useful in privatization settings with entities such as names or locations. For \textsc{GLiNER}, we use the pretrained \textsc{urchade/gliner\_multi\_pii-v1} model, provided by the authors of the original work. As with Presidio, we also test both \textit{redact} and \textit{replace}.

\paragraph{\textsc{SanText}.}
\textsc{SanText} \cite{yue-etal-2021-differential} is a word-level metric Differential Privacy (DP) obfuscation mechanism, which performs word-by-word replacements to provide private outputs with DP guarantees. An important parameter is the privacy budget ($\varepsilon$), which we set for all DP mechanisms to be on the \textit{document-level}. For each dataset, we set the dataset-specific document-level budget to be $\varepsilon\cdot avg. \ tokens$ (as per \textsc{nltk}), for $\varepsilon \in \{0.5, 1\}$.

\paragraph{\textsc{DP-MLM}.}
\textsc{DP-MLM} \cite{meisenbacher-etal-2024-dp} performs contextualized token replacements for DP token outputs, rewriting whole texts sequentially using Masked Language Models. In a similar manner to \textsc{SanText}, we set base $\varepsilon \in \{10, 25\}$ (following the original work), which are scaled by the averaged token count per text in a dataset. We use the original public implementation, which is built on a \textsc{RoBERTa-base} model.

\paragraph{\textsc{DP-BART}.}
The final DP-based mechanism, \textsc{DP-BART} \cite{igamberdiev-habernal-2023-dp}, rewrites text documents under DP guarantees, using a encoder-decoder (by default \textsc{BART}) model. Following the original work, we use document-level $\varepsilon \in \{1000, 2000\}$ with a \textsc{BART-large} model.

\paragraph{\textsc{RUPTA}.}
\textsc{RUPTA} \cite{yang-etal-2025-robust}, or \textit{Robust Utility-Preserving Text Anonymization}, is a LLM-based anonymization method which splits the task of text anonymization into three components: a privacy evaluator, a utility evaluator, and an optimization component. We utilize this framework with a \textsc{Qwen2.5-7B-Instruct-GPTQ-Int4} model, with otherwise default parameters and prompts.

\paragraph{Privacy Filter.}
OpenAI's \textit{Privacy Filter}\footnote{\scriptsize\url{https://huggingface.co/openai/privacy-filter}} is an open-source NER tool that specializes in detecting PII and masking this information. We use the default \textsc{openai/privacy-filter} model, and implement a similar \textit{redact} or \textit{replace} setup as before.

\subsection{Evaluation Procedure}
For each privatized dataset counterpart (i.e., for Reddit-25, Reddit-50, and Twitter-10), we fine-tune a new HSD classification model, in the same manner as in Section \ref{sec:foundations}, but where the original subset in question is replaced by the privatized version (both train and test set). The micro-F1 scores on the test set are recorded to represent $H_p$. 

The trained models are then used to obtain the probe accuracies and FPR $\sigma$ scores. For each privatized dataset, we repeat the training and scoring procedure three times (on different shuffles of the training sets, seeds 41-43), to account for variances in model optimization. The average scores of these three runs are presented as the final scores. Following this, we retain only the 20\% (seed: 42) test splits, as before, and measure adversarial micro-F1 using the corresponding adversarial classification models introduced in Section \ref{sec:foundations}. Together, these three scores represent the $m_p$ components, which comprise the $\Delta Privacy$ when averaged.

Combining all of the above scores and achieving the $\Delta$ results using baseline values, we then are able to calculate the $PrivHSD$ trade-offs for each privatization setting. Although not included in the trade-off, we also measure text coherence (i.e., quality) via perplexity, measured using a GPT-2 model \cite{radford2019language}. The complete results and score breakdowns are presented in Table \ref{tab:results}.

\section{Towards Privacy-preserving HSD}
Intuitively, and as evidenced by the results of Table \ref{tab:results}, all tested text privatization mechanisms are generalized, i.e., not tailored for utility preservation in the context of HSD. As such, their demonstrated strong ability in protecting privacy often comes at the cost of severely diminished HSD performance. In response to this, we design a privacy-preserving and author-agnostic text privatization method, with the goal of optimizing the $PrivHSD$ trade-off.

\subsection{The \textsc{AgnoSpeech} Method}
The core of \textsc{AgnoSpeech} is simple: a privacy-preserving HSD method should ideally redact all signals not immediately necessary for identifying hate speech, while keeping these hate speech cues intact. We also optimize for readability, motivated by improving methods that redact without replacement, or alternatively, that leave texts largely incoherent (as with stricter DP-based privatization).

\begin{table*}[t]
\centering
\resizebox{\linewidth}{!}{
\begin{tabular}{l | ccccccc | ccccccc | ccccccc | c}
\hline
 & \multicolumn{7}{c|}{\textbf{Reddit-25}} & \multicolumn{7}{c|}{\textbf{Reddit-50}} & \multicolumn{7}{c|}{\textbf{Twitter-10}} \\ \cline{2-22}
\textbf{Method} & PPL $\downarrow$ & HSD $\uparrow$ & Probe $\downarrow$ & $\eta^2$ $\downarrow$ & FPR $\sigma$ $\downarrow$ & Adv $\downarrow$ & TO $\uparrow$ & PPL $\downarrow$ & HSD $\uparrow$ & Probe $\downarrow$ & $\eta^2$ $\downarrow$ & FPR $\sigma$ $\downarrow$ & Adv $\downarrow$ & TO $\uparrow$ & PPL $\downarrow$ & HSD $\uparrow$ & Probe $\downarrow$ & $\eta^2$ $\downarrow$ & FPR $\sigma$ $\downarrow$ & Adv $\downarrow$ & TO $\uparrow$ & $\overline{TO}$ $\uparrow$ \\
\hline
\textit{Baseline} & 377 & 89.0\% & 39.8\% & 0.22 & 0.11 & 19.0\% & -- & 346 & 86.0\% & 21.7\% & 0.22 & 0.16 & 13.1\% & -- & 250 & 91.0\% & 88.2\% & 0.61 & 0.21 & 87.2\% & -- & -- \\
\hline
Presidio (redact) & 8034 & 90.0\% & 40.5\% & 0.21 & 0.00 & 5.6\% & 0.44 & 5298 & 86.8\% & 21.4\% & 0.23 & 0.00 & 0.6\% & 0.50 & 1050 & 87.2\% & 82.6\% & 0.51 & 0.01 & 10.7\% & 0.34 & 0.43 \\
Presidio (replace) & 376 & 90.2\% & 39.0\% & 0.21 & 0.00 & 4.8\% & 0.46 & 344 & 86.9\% & 19.2\% & 0.23 & 0.00 & 0.8\% & 0.51 & 273 & 87.6\% & 83.5\% & 0.50 & 0.02 & 11.2\% & 0.34 & 0.44 \\
\textsc{GLiNER} (redact) & 467 & 87.7\% & 34.3\% & 0.20 & 0.00 & 4.3\% & 0.49 & 436 & 84.9\% & 19.7\% & 0.19 & 0.00 & 1.1\% & 0.52 & 556 & 86.6\% & 81.4\% & 0.50 & 0.00 & 2.4\% & 0.35 & 0.45 \\
\textsc{GLiNER} (replace) & 347 & 87.4\% & 36.7\% & 0.19 & 0.00 & 5.6\% & 0.45 & 317 & 83.7\% & 20.6\% & 0.19 & 0.00 & 1.1\% & 0.49 & 167 & 87.1\% & 83.6\% & 0.53 & 0.01 & 3.6\% & 0.34 & 0.43 \\
\textsc{SanText} ($\varepsilon = 0.5$) & 15179 & 69.7\% & 8.7\% & 0.11 & 0.04 & 4.8\% & 0.40 & 14346 & 75.3\% & 5.8\% & 0.14 & 0.01 & 2.2\% & 0.56 & 27953 & 69.7\% & 46.6\% & 0.05 & 0.00 & 0.4\% & -0.15 & 0.27 \\
\textsc{SanText} ($\varepsilon = 1$) & 10594 & 77.6\% & 11.8\% & 0.13 & 0.04 & 4.8\% & 0.47 & 9983 & 77.3\% & 7.3\% & 0.17 & 0.00 & 1.4\% & 0.58 & 28473 & 69.7\% & 49.4\% & 0.05 & 0.00 & 0.7\% & -0.16 & 0.30 \\
\textsc{DP-MLM} ($\varepsilon = 10$) & 3382 & 73.4\% & 18.8\% & 0.16 & 0.02 & 3.9\% & 0.39 & 3297 & 74.8\% & 11.4\% & 0.20 & 0.03 & 1.7\% & 0.42 & 8275 & 79.7\% & 69.5\% & 0.30 & 0.06 & 5.1\% & 0.07 & 0.29 \\
\textsc{DP-BART} ($\varepsilon = 25$) & 1907 & 75.3\% & 24.2\% & 0.15 & 0.03 & 3.5\% & 0.38 & 2040 & 77.9\% & 10.7\% & 0.22 & 0.01 & 1.1\% & 0.48 & 4621 & 82.2\% & 71.2\% & 0.36 & 0.00 & 7.8\% & 0.22 & 0.36 \\
\textsc{DP-BART} ($\varepsilon = 1000$) & 7859 & 70.0\% & 4.8\% & 0.14 & 0.05 & 6.5\% & 0.35 & 5010 & 71.9\% & 3.3\% & 0.22 & 0.02 & 1.7\% & 0.46 & 5543 & 69.6\% & 48.5\% & 0.02 & 0.00 & 1.8\% & -0.15 & 0.22 \\
\textsc{DP-BART} ($\varepsilon = 2000$) & 2475 & 69.0\% & 5.9\% & 0.24 & 0.04 & 5.2\% & 0.26 & 821 & 72.9\% & 3.2\% & 0.23 & 0.04 & 1.7\% & 0.43 & 485 & 69.7\% & 47.3\% & 0.03 & 0.00 & 1.3\% & -0.15 & 0.18 \\
RUPTA & 340 & 80.7\% & 32.8\% & 0.14 & 0.00 & 6.5\% & 0.43 & 345 & 76.8\% & 19.6\% & 0.18 & 0.00 & 1.7\% & 0.42 & 237 & 87.3\% & 83.7\% & 0.49 & 0.03 & 11.7\% & 0.32 & 0.39 \\
Privacy Filter (redact) & 410 & 88.9\% & 37.1\% & 0.20 & 0.00 & 6.1\% & 0.45 & 363 & 86.4\% & 21.5\% & 0.23 & 0.00 & 1.1\% & 0.47 & 594 & 86.2\% & 81.5\% & 0.50 & 0.02 & 3.9\% & 0.31 & 0.41 \\
Privacy Filter (replace) & 354 & 89.0\% & 36.7\% & 0.20 & 0.00 & 6.5\% & 0.45 & 300 & 86.1\% & 21.5\% & 0.23 & 0.00 & 1.1\% & 0.47 & 86 & 87.5\% & 84.2\% & 0.49 & 0.02 & 3.5\% & \underline{0.37} & 0.43 \\
\hline \hline
\textsc{AgnoSpeech} L1 (fast) & 393 & 90.2\% & 38.7\% & 0.20 & 0.00 & 5.6\% & 0.47 & 359 & 86.6\% & 19.7\% & 0.23 & 0.00 & 0.8\% & 0.49 & 158 & 88.7\% & 84.5\% & 0.56 & 0.02 & 3.1\% & \textbf{0.39} & 0.45 \\
\textsc{AgnoSpeech} L1 (per.) & 281 & 81.4\% & 32.9\% & 0.15 & 0.00 & 4.8\% & 0.45 & 265 & 81.5\% & 19.8\% & 0.18 & 0.01 & 0.6\% & 0.49 & 132 & 86.9\% & 82.1\% & 0.51 & 0.04 & 1.9\% & 0.31 & 0.42 \\
\textsc{AgnoSpeech} L2 (fast) & 3603 & 89.6\% & 16.7\% & 0.20 & 0.00 & 6.5\% & 0.59 & 3309 & 87.4\% & 10.5\% & 0.22 & 0.00 & 1.9\% & \textbf{0.61} & 1556 & 78.7\% & 70.3\% & 0.33 & 0.12 & 0.6\% & -0.06 & 0.38 \\
\textsc{AgnoSpeech} L2 (per.) & 4215 & 87.3\% & 16.9\% & 0.16 & 0.00 & 5.6\% & \textbf{0.61} & 3941 & 84.9\% & 12.3\% & 0.22 & 0.00 & 1.4\% & 0.57 & 3503 & 83.9\% & 72.9\% & 0.39 & 0.00 & 1.3\% & 0.30 & \underline{0.49} \\
\textsc{AgnoSpeech} L3 (fast) & 1329 & 88.7\% & 26.0\% & 0.20 & 0.00 & 3.9\% & 0.55 & 1288 & 88.4\% & 12.6\% & 0.22 & 0.00 & 2.5\% & \underline{0.59} & 632 & 86.1\% & 75.1\% & 0.43 & 0.01 & 0.7\% & \underline{0.37} & \textbf{0.50} \\
\textsc{AgnoSpeech} L3 (per.) & 1213 & 89.3\% & 22.7\% & 0.19 & 0.00 & 3.5\% & \underline{0.60} & 1127 & 86.8\% & 14.8\% & 0.21 & 0.00 & 1.9\% & 0.56 & 708 & 85.6\% & 75.1\% & 0.43 & 0.01 & 0.7\% & 0.34 & \textbf{0.50} \\
\hline
\end{tabular}
}
\caption{Complete results of the case study (top) and \textsc{AgnoSpeech} evaluations (bottom). \textit{PPL} denotes perplexity (coherence) and \textit{HSD} is hate speech classification performance (in micro-F1). The four remaining privacy metrics are those introduced in Section \ref{sec:foundations} and Table \ref{tab:probe}. The \textit{TO} score denotes the $PrivHSD$ score (Equation \ref{eq:to}), or relative gains in the four privacy metrics over \textit{HSD} preservation. $\uparrow$ denotes higher is better, and $\downarrow$ means lower is better. The best and second-best average trade-offs ($TO$) per dataset and overall ($\overline{TO}$) are \textbf{bolded} and \underline{underlined}, respectively. We note that while $TO$ incorporates the normalized HSD results (Equation \ref{eq:to}), $HSD$ presents the raw F1 scores.}
\label{tab:results}
\end{table*}

\begin{figure}[t]
    \centering
    \includegraphics[width=0.95\linewidth]{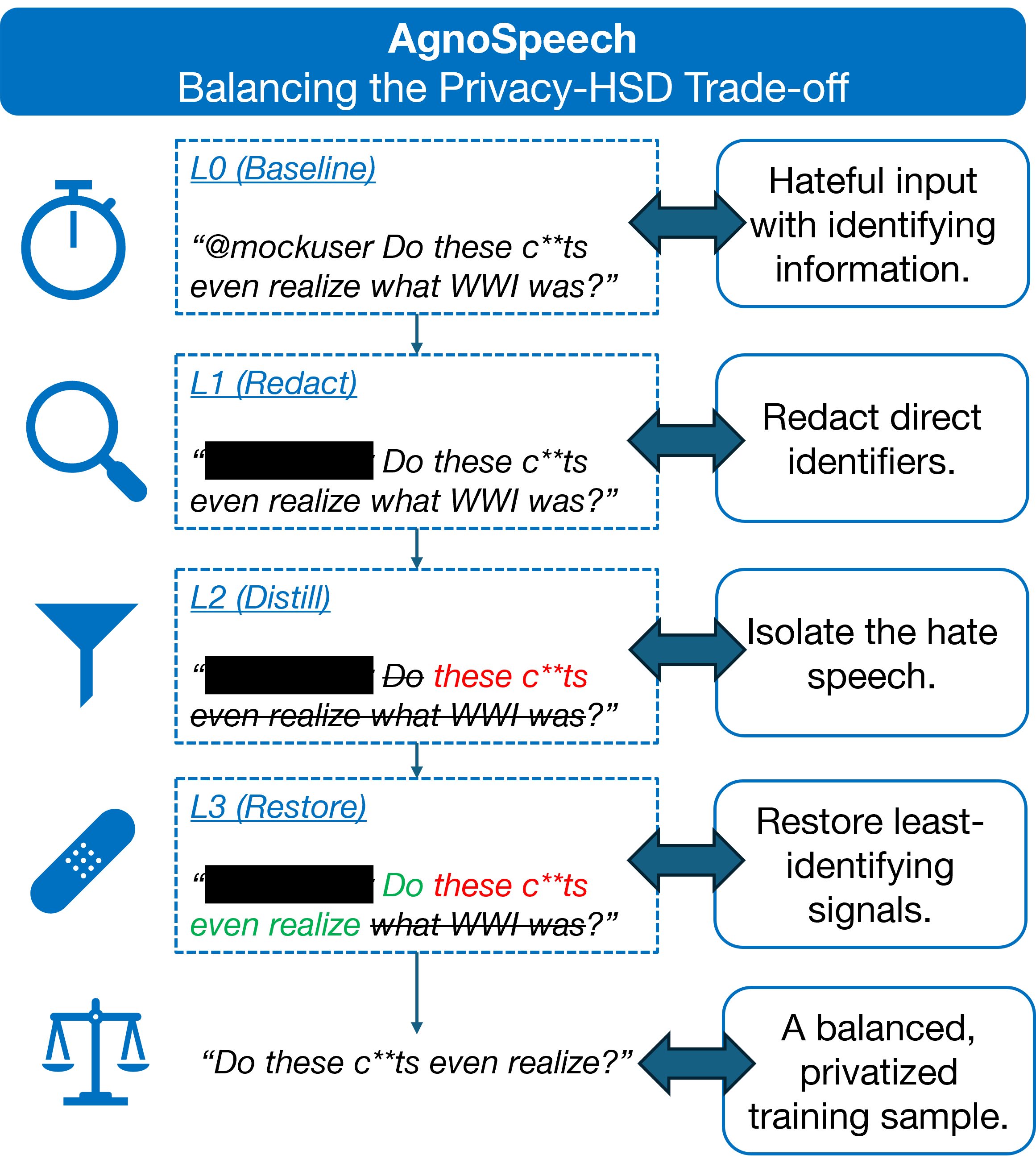}
    \caption{An overview of AgnoSpeech, a multi-tiered method for facilitating HSD, but not at the cost of privacy. Note: harmful language has been pre-redacted.}
    \label{fig:agnospeech}
\end{figure}

We implemented \textsc{AgnoSpeech} as a three-tiered method, in which each level addresses a separate concern and transforms an input text accordingly. Additionally, we prepare two variants of \textsc{AgnoSpeech}: (1) \textit{fast}, which is built for efficiency and throughput, and (2) \textit{performance}, which is built using more computationally expensive models. While we introduce each level below (and illustratively in Figure \ref{fig:agnospeech}), implementation specifics can be found in our provided code repository.

\paragraph{L0.} The un-privatized input is denoted as L0.

\paragraph{L1.} In the L1 (\textit{Redact}) level, we take motivation from entity-based anonymization tools, where PII and other direct identifiers are immediately removed, as these intuitively do not contribute to hate speech signals. In the \textit{fast} variant, redaction is accomplished via a simple regex detector, whereas the \textit{performance} model augments regex detection with Microsoft Presidio's analyzer tool.

\paragraph{L2.} Building on our findings from Table \ref{tab:probe}, L2 (\textit{Distill}) uses linear attribution (\textit{fast}) or token saliency (\textit{performance}) to learn hate speech signal importance from a trained proxy model. In the linear attribution component, a logistic regression model is trained on joint unigram and bigram vocabulary, thus producing a coefficient for each member of this joint vocabulary. During lookup, each weight (regression coefficient) is multiplied by the corpus TF-IDF weight, producing a final importance score per word. Finally, only a fraction of the top-scoring words (i.e., those most important to detecting hate speech, default: 60\%) are kept. 

The \textit{performance} variant improves upon this distillation approach by measuring the \textit{saliency} of each word by calculating the change in regressor prediction confidence (probability) after removing the word from the input text. This is done sequentially for each word in a given text, and subsequently, a similar percentage filtering is performed, with the same default 60\%, i.e., to remove the bottom 40\% of tokens not important to HSD.

\paragraph{L3.} Left as an optional third transformation following L2, L3 (\textit{Restore}) re-introduces pruned tokens selectively and randomly, with the motivation that sweepingly removing all non-hate tokens can severely degrade readability and coherence. This module, identical for both \textit{fast} and \textit{performance}, takes an \textit{intensity} input parameter ($i \in [0, 1]$), and for each removed token from L2, a random number in that range is calculated, and if $< i$, the word is restored to the final text. While this re-introduces the risk of restoring author-identifying signals, it supports in the balance of redaction and readability.

\paragraph{Computational costs.}
Both the \textit{fast} and \textit{performance} variants of \textsc{AgnoSpeech} are highly lightweight and performant. We confirm this in runtime benchmarks on Reddit-25, presented in Table \ref{tab:speed}, which show that even with one-time setup costs and running the full L1--3 pipeline, average runtime per text is small and thus scalable.

\section{Analysis and Discussion}
We reflect on our findings, focusing on the implications for privacy-preserving HSD going forward.

\paragraph{Results and comparative analysis.}
Table \ref{tab:results} contains the complete results of our evaluations of \textsc{AgnoSpeech} along with the selected comparative methods. To test \textsc{AgnoSpeech} comprehensively, we not only evaluate the \textit{fast} and \textit{performance} variants but run an ablation by iteratively testing each tier, i.e., L1 only, L1 and L2, and L1--3. These are denoted in Table \ref{tab:results} by the highest included tier.

The results consistently show that \textsc{AgnoSpeech} achieves positive and competitive $PrivHSD$ trade-offs, which can be attributed to strong privacy preservation while also preserving the HSD signal. This comes in contrast to methods such as \textsc{DP-BART} and \textsc{SanText} results, which may achieve relatively high privacy scores, but at the cost of extreme coherence and utility loss (see the $PPL$ and $HSD$ scores). \textsc{AgnoSpeech} L2 achieves high and competitive trade-offs at lower perplexity, and furthermore, L3 corrects this coherence loss with minimal trade-off degradation.

When purely comparing HSD performance to $PrivHSD$ trade-offs, \textsc{AgnoSpeech} excels. All L1-3 variants generally maintain HSD classification performance to the level of methods such as Presidio, \textsc{GLiNER}, and Privacy Filter, while still achieving higher trade-offs. This showcases the domain-specific focus of \textsc{AgnoSpeech}, i.e., maintaining HSD but in a privacy-preserving manner.

A final comparison is the effectiveness between the \textit{fast} and \textit{performance} variants of our method. As one can see, the \textit{performance} variant usually slightly sacrifices HSD results but rectifies this with stronger privacy preservation, thus often achieving higher trade-offs. This makes a case for the more computationally heavy variant of \textsc{AgnoSpeech}, but the results also show that \textit{fast} is an acceptable option in resource-constrained environments.

\paragraph{The need for a balance.}
The empirical results of our case study and \textsc{AgnoSpeech} evaluations teach an important lesson, in that achieving \say{acceptable} trade-offs in $PrivHSD$ is not necessarily about pushing privacy preservation to the extreme (as with some DP-based methods). In doing so, one end of the trade-off might be optimized, but this comes with undesirable consequences, e.g, up to 20 percentage points of HSD performance loss.

Going forward, this points to potential improvements in quantifying $PrivHSD$, such as incorporating perplexity or other quality metrics as an aspect of \textit{utility preservation}. In the same vein, we do not consider semantic similarity or grammatical correctness, two examples of important criteria for producing privatized, but usable datasets for HSD.

\begin{table}[t]
\centering
\resizebox{\linewidth}{!}{
\begin{tabular}{l|ccc}
\hline
\multicolumn{1}{c|}{\textbf{Setting}} & \multicolumn{1}{c}{\textbf{HSD (s)}} & \multicolumn{1}{c}{\textbf{Build (s)}} & \multicolumn{1}{c}{\textbf{Priv. (ms/post)}} \\
\hline
\textit{fast}, CPU & 0.76 & 3.19 & 0.45 \\
\textit{performance}, GPU & 23.5 & 10.4 & 7.5
\end{tabular}
}
\caption{Runtime benchmarks for the full \textsc{AgnoSpeech} pipeline on Reddit-25. Runtime includes \textbf{HSD} model download / fitting, L1--3 \textbf{build}ing, and text \textbf{priv}atization. The \textit{fast} variant is run on a two-core Colab vCPU, and \textit{performance} is run on a single Colab T4 GPU. Times are in seconds (s) or milliseconds (ms).}
\label{tab:speed}
\end{table}

\paragraph{Negative results and areas for improvement.}
While several methods, particularly \textsc{AgnoSpeech}, are successful in addressing probe accuracy and adversarial F1, most struggle with author-attributable variance. This is especially pronounced in the Reddit datasets. Not mitigating these risks propagates the potential of encoding authorship, and this also carries fairness implications for HSD systems.

Similarly, while we observe considerable success in addressing probe accuracy (i.e., mitigating authorship entanglement), this intertwining is never completely diminished, especially in the case of Twitter. This also harks back to our baseline tests in the linear probing experiments, which show that even the non-fine-tuned models encode authorship information, a challenge which the current version of \textsc{Agnospeech} may not be currently equipped to fully tackle. Devising methods to separate these signals completely and to mitigate inherent privacy vulnerabilities, while also maintaining HSD performance, remains a crucial point for future research.

\section{Conclusion}
We establish the foundations of the privacy-HSD trade-off, motivated by the important need for combating online hate speech while also preserving the right to privacy. Grounded in a set of metrics that measure the author-identifiability in HSD models, we benchmark existing text privatization methods against our proposed \textsc{AgnoSpeech}, highlighting the complex relationship in balancing privacy and HSD, but also demonstrating the merits of tailored methods such as \textsc{AgnoSpeech} in doing so. Our work opens the path for privacy-preserving HSD, while also emphasizing key areas for improvement in the design and evaluation of methods therein. 

\section*{Limitations}
The primary limitation of our work comes with the assumption of constrained author sets, which allowed for controlled experiments. We argue, however, that this still presents a real-world use case, where either an internal malicious actor (who has access to raw HSD datasets) or a capable adversary could plausibly filter down datasets by frequently writing users, for example within a subreddit.

Another limitation is the dataset dependence of \textsc{AgnoSpeech}, particularly in L2 where the saliency measurements must be derived from a model trained on selected data. We demonstrated cross-dataset generalizability by basing L2 in Reddit and showing that \textsc{AgnoSpeech} still remains effective on Twitter. Nevertheless, future work should seek to improve this generalizability.

For the probing study and evaluations, we rely solely on \textsc{bert-base-cased} as the underlying proxy. While we believe this choice is justified as this particular model is the bedrock for many similar ones thereafter, we suggest further evaluations be conducted on diverse model architectures.

We target author re-identification, which is one form of privacy risk that an adversary may exploit. Others, such as attribute inference (e.g., demographics or political affiliation), are not considered but would constitute logical extensions of our work.

A final limitation pertains to the data itself, in that the two datasets we use are well established but not up-to-date (the Reddit dataset was published in 2021 and the Twitter data is from 2016). Follow-up work should seek to evaluate privacy-preserving HSD methods on newer datasets, if available, in order to maintain practical usability in today's world.

\section*{Ethical Considerations}
Our work operates at the intersection of combating online hate speech and protecting the human right to privacy. We show the ethical complexities therein, specifically in the danger that creating high-performant HSD tools might inadvertently entangle authorship information, thereby enabling secondary, unintended uses such as profiling or re-identification tools. Our method, \textsc{AgnoSpeech}, begins the discussion on how these two ethical concerns can be addressed in tandem and in balance.

We confirm that the datasets in this work were used in line with their intended purpose (HSD) and according to their license of use (CC BY-NC 4.0 for Reddit and CC BY 4.0 for Twitter metadata).

We prevented harm by converting author usernames to pseudonymous IDs, i.e., integers. We warn that the utilized datasets contain hateful and distressing language, and thus advise caution.

\section*{Acknowledgments}
We thank the Council of Europe and New Democratic Pact for Europe for hosting the \textit{Hack the Hate, Renew Democracy} hackathon, which grounded this work and brought the research team together.

\bibliography{custom}

@article{gandhi2024hate,
  title={Hate speech detection: A comprehensive review of recent works},
  author={Gandhi, Ankita and Ahir, Param and Adhvaryu, Kinjal and Shah, Pooja and Lohiya, Ritika and Cambria, Erik and Poria, Soujanya and Hussain, Amir},
  journal={Expert Systems},
  volume={41},
  number={8},
  pages={e13562},
  year={2024},
  publisher={Wiley Online Library},
  doi={10.1111/exsy.13562}
}

@article{alkomah2022literature,
  title={A literature review of textual hate speech detection methods and datasets},
  author={Alkomah, Fatimah and Ma, Xiaogang},
  journal={Information},
  volume={13},
  number={6},
  pages={273},
  year={2022},
  publisher={MDPI},
  doi={10.3390/info13060273}
}

@article{10.1145/3232676,
author = {Fortuna, Paula and Nunes, S\'{e}rgio},
title = {A Survey on Automatic Detection of Hate Speech in Text},
year = {2018},
issue_date = {July 2019},
publisher = {Association for Computing Machinery},
address = {New York, NY, USA},
volume = {51},
number = {4},
issn = {0360-0300},
url = {https://doi.org/10.1145/3232676},
doi = {10.1145/3232676},
journal = {ACM Comput. Surv.},
month = jul,
articleno = {85},
numpages = {30}
}

@ARTICLE{10848067,
  author={Albladi, Aish and Islam, Minarul and Das, Amit and Bigonah, Maryam and Zhang, Zheng and Jamshidi, Fatemeh and Rahgouy, Mostafa and Raychawdhary, Nilanjana and Marghitu, Daniela and Seals, Cheryl},
  journal={IEEE Access}, 
  title={Hate Speech Detection Using Large Language Models: A Comprehensive Review}, 
  year={2025},
  volume={13},
  number={},
  pages={20871-20892},
  doi={10.1109/ACCESS.2025.3532397}}

@article{rawat2024hate,
  title={Hate speech detection in social media: Techniques, recent trends, and future challenges},
  author={Rawat, Anchal and Kumar, Santosh and Samant, Surender Singh},
  journal={Wiley Interdisciplinary Reviews: Computational Statistics},
  volume={16},
  number={2},
  pages={e1648},
  year={2024},
  publisher={Wiley Online Library},
  doi={10.1002/wics.1648}
}

@ARTICLE{10025718,
  author={Mansur, Zainab and Omar, Nazlia and Tiun, Sabrina},
  journal={IEEE Access}, 
  title={Twitter Hate Speech Detection: A Systematic Review of Methods, Taxonomy Analysis, Challenges, and Opportunities}, 
  year={2023},
  volume={11},
  number={},
  pages={16226-16249},
  doi={10.1109/ACCESS.2023.3239375}}

@article{li2012theories,
  title={Theories in online information privacy research: A critical review and an integrated framework},
  author={Li, Yuan},
  journal={Decision support systems},
  volume={54},
  number={1},
  pages={471--481},
  year={2012},
  publisher={Elsevier},
  doi={10.1016/j.dss.2012.06.010}
}

@article{slavkovic2023statistical,
  title={Statistical data privacy: A song of privacy and utility},
  author={Slavkovi{\'c}, Aleksandra and Seeman, Jeremy},
  journal={Annual Review of Statistics and Its Application},
  volume={10},
  number={1},
  pages={189--218},
  year={2023},
  publisher={Annual Reviews},
  doi={10.1146/annurev-statistics-033121-112921}
}

@inproceedings{schmidt-wiegand-2017-survey,
    title = "A Survey on Hate Speech Detection using Natural Language Processing",
    author = "Schmidt, Anna  and
      Wiegand, Michael",
    editor = "Ku, Lun-Wei  and
      Li, Cheng-Te",
    booktitle = "Proceedings of the Fifth International Workshop on Natural Language Processing for Social Media",
    month = apr,
    year = "2017",
    address = "Valencia, Spain",
    publisher = "Association for Computational Linguistics",
    url = "https://aclanthology.org/W17-1101/",
    doi = "10.18653/v1/W17-1101",
    pages = "1--10"
}

@article{malik2025deep,
  title={Deep learning for hate speech detection: a comparative study},
  author={Malik, Jitendra Singh and Qiao, Hezhe and Pang, Guansong and van den Hengel, Anton},
  journal={International Journal of Data Science and Analytics},
  volume={20},
  number={4},
  pages={3053--3068},
  year={2025},
  publisher={Springer},
  doi={10.1007/s41060-024-00650-6}
}

@article{10.1145/3583067,
author = {Govers, Jarod and Feldman, Philip and Dant, Aaron and Patros, Panos},
title = {Down the Rabbit Hole: Detecting Online Extremism, Radicalisation, and Politicised Hate Speech},
year = {2023},
issue_date = {December 2023},
publisher = {Association for Computing Machinery},
address = {New York, NY, USA},
volume = {55},
number = {14s},
issn = {0360-0300},
url = {https://doi.org/10.1145/3583067},
doi = {10.1145/3583067},
journal = {ACM Comput. Surv.},
month = jul,
articleno = {319},
numpages = {35}
}

@article{gitari2015lexicon,
  title={A Lexicon-based Approach for Hate Speech Detection},
  author={Gitari, Njagi Dennis and Zuping, Zhang and Damien, Hanyurwimfura and Long, Jun},
  journal={International Journal of Multimedia and Ubiquitous Engineering},
  volume={10},
  number={4},
  pages={215--230},
  year={2015},
  doi={10.14257/ijmue.2015.10.4.21}
}

@inproceedings{davidson2017automated,
  title={Automated hate speech detection and the problem of offensive language},
  author={Davidson, Thomas and Warmsley, Dana and Macy, Michael and Weber, Ingmar},
  booktitle={Proceedings of the international AAAI conference on web and social media},
  volume={11},
  pages={512--515},
  year={2017},
  doi={10.1609/icwsm.v11i1.14955}
}

@inproceedings{10.1145/2740908.2742760,
author = {Djuric, Nemanja and Zhou, Jing and Morris, Robin and Grbovic, Mihajlo and Radosavljevic, Vladan and Bhamidipati, Narayan},
title = {Hate Speech Detection with Comment Embeddings},
year = {2015},
isbn = {9781450334730},
publisher = {Association for Computing Machinery},
address = {New York, NY, USA},
url = {https://doi.org/10.1145/2740908.2742760},
doi = {10.1145/2740908.2742760},
booktitle = {Proceedings of the 24th International Conference on World Wide Web},
pages = {29–30},
numpages = {2},
location = {Florence, Italy},
series = {WWW '15 Companion}
}

@inproceedings{10.1145/2872427.2883062,
author = {Nobata, Chikashi and Tetreault, Joel and Thomas, Achint and Mehdad, Yashar and Chang, Yi},
title = {Abusive Language Detection in Online User Content},
year = {2016},
isbn = {9781450341431},
publisher = {International World Wide Web Conferences Steering Committee},
address = {Republic and Canton of Geneva, CHE},
url = {https://doi.org/10.1145/2872427.2883062},
doi = {10.1145/2872427.2883062},
booktitle = {Proceedings of the 25th International Conference on World Wide Web},
pages = {145–153},
numpages = {9},
location = {Montr\'{e}al, Qu\'{e}bec, Canada},
series = {WWW '16}
}

@inproceedings{park-fung-2017-one,
    title = "One-step and Two-step Classification for Abusive Language Detection on {T}witter",
    author = "Park, Ji Ho  and
      Fung, Pascale",
    editor = "Waseem, Zeerak  and
      Chung, Wendy Hui Kyong  and
      Hovy, Dirk  and
      Tetreault, Joel",
    booktitle = "Proceedings of the First Workshop on Abusive Language Online",
    month = aug,
    year = "2017",
    address = "Vancouver, BC, Canada",
    publisher = "Association for Computational Linguistics",
    url = "https://aclanthology.org/W17-3006/",
    doi = "10.18653/v1/W17-3006",
    pages = "41--45"
}

@ARTICLE{9139953,
  author={Zhou, Yanling and Yang, Yanyan and Liu, Han and Liu, Xiufeng and Savage, Nick},
  journal={IEEE Access}, 
  title={Deep Learning Based Fusion Approach for Hate Speech Detection}, 
  year={2020},
  volume={8},
  number={},
  pages={128923-128929},
  doi={10.1109/ACCESS.2020.3009244}}

@inproceedings{10.1145/3357384.3358040,
author = {Rizos, Georgios and Hemker, Konstantin and Schuller, Bj\"{o}rn},
title = {Augment to Prevent: Short-Text Data Augmentation in Deep Learning for Hate-Speech Classification},
year = {2019},
isbn = {9781450369763},
publisher = {Association for Computing Machinery},
address = {New York, NY, USA},
url = {https://doi.org/10.1145/3357384.3358040},
doi = {10.1145/3357384.3358040},
booktitle = {Proceedings of the 28th ACM International Conference on Information and Knowledge Management},
pages = {991–1000},
numpages = {10},
location = {Beijing, China},
series = {CIKM '19}
}

@inproceedings{mozafari2019bert,
  title={A BERT-based transfer learning approach for hate speech detection in online social media},
  author={Mozafari, Marzieh and Farahbakhsh, Reza and Crespi, Noel},
  booktitle={International conference on complex networks and their applications},
  pages={928--940},
  year={2019},
  organization={Springer},
  doi={10.1007/978-3-030-36687-2_77}
}

@inproceedings{ghorbanpour-etal-2025-prompting,
    title = "Can Prompting {LLM}s Unlock Hate Speech Detection across Languages? A Zero-shot and Few-shot Study",
    author = "Ghorbanpour, Faeze  and
      Dementieva, Daryna  and
      Fraser, Alexander",
    editor = "Calabrese, Agostina  and
      de Kock, Christine  and
      Nozza, Debora  and
      Plaza-del-Arco, Flor Miriam  and
      Talat, Zeerak  and
      Vargas, Francielle",
    booktitle = "Proceedings of the The 9th Workshop on Online Abuse and Harms (WOAH)",
    month = aug,
    year = "2025",
    address = "Vienna, Austria",
    publisher = "Association for Computational Linguistics",
    url = "https://aclanthology.org/2025.woah-1.39/",
    pages = "413--425",
    ISBN = "979-8-89176-105-6"
}

@INPROCEEDINGS{10459901,
  author={Guo, Keyan and Hu, Alexander and Mu, Jaden and Shi, Ziheng and Zhao, Ziming and Vishwamitra, Nishant and Hu, Hongxin},
  booktitle={2023 International Conference on Machine Learning and Applications (ICMLA)}, 
  title={An Investigation of Large Language Models for Real-World Hate Speech Detection}, 
  year={2023},
  volume={},
  number={},
  pages={1568-1573},
  doi={10.1109/ICMLA58977.2023.00237}}

@inproceedings{toraman-etal-2022-large,
    title = "Large-Scale Hate Speech Detection with Cross-Domain Transfer",
    author = "Toraman, Cagri  and
      {\c{S}}ahinu{\c{c}}, Furkan  and
      Yilmaz, Eyup",
    editor = "Calzolari, Nicoletta  and
      B{\'e}chet, Fr{\'e}d{\'e}ric  and
      Blache, Philippe  and
      Choukri, Khalid  and
      Cieri, Christopher  and
      Declerck, Thierry  and
      Goggi, Sara  and
      Isahara, Hitoshi  and
      Maegaard, Bente  and
      Mariani, Joseph  and
      Mazo, H{\'e}l{\`e}ne  and
      Odijk, Jan  and
      Piperidis, Stelios",
    booktitle = "Proceedings of the Thirteenth Language Resources and Evaluation Conference",
    month = jun,
    year = "2022",
    address = "Marseille, France",
    publisher = "European Language Resources Association",
    url = "https://aclanthology.org/2022.lrec-1.238/",
    pages = "2215--2225"
}

@inproceedings{yang-etal-2023-hare,
    title = "{HARE}: Explainable Hate Speech Detection with Step-by-Step Reasoning",
    author = "Yang, Yongjin  and
      Kim, Joonkee  and
      Kim, Yujin  and
      Ho, Namgyu  and
      Thorne, James  and
      Yun, Se-Young",
    editor = "Bouamor, Houda  and
      Pino, Juan  and
      Bali, Kalika",
    booktitle = "Findings of the Association for Computational Linguistics: EMNLP 2023",
    month = dec,
    year = "2023",
    address = "Singapore",
    publisher = "Association for Computational Linguistics",
    url = "https://aclanthology.org/2023.findings-emnlp.365/",
    doi = "10.18653/v1/2023.findings-emnlp.365",
    pages = "5490--5505"
}

@inproceedings{rottger-etal-2021-hatecheck,
    title = "{H}ate{C}heck: Functional Tests for Hate Speech Detection Models",
    author = {R{\"o}ttger, Paul  and
      Vidgen, Bertie  and
      Nguyen, Dong  and
      Waseem, Zeerak  and
      Margetts, Helen  and
      Pierrehumbert, Janet},
    editor = "Zong, Chengqing  and
      Xia, Fei  and
      Li, Wenjie  and
      Navigli, Roberto",
    booktitle = "Proceedings of the 59th Annual Meeting of the Association for Computational Linguistics and the 11th International Joint Conference on Natural Language Processing (Volume 1: Long Papers)",
    month = aug,
    year = "2021",
    address = "Online",
    publisher = "Association for Computational Linguistics",
    url = "https://aclanthology.org/2021.acl-long.4/",
    doi = "10.18653/v1/2021.acl-long.4",
    pages = "41--58"
}

@inproceedings{mathew2021hatexplain,
  title={Hatexplain: A benchmark dataset for explainable hate speech detection},
  author={Mathew, Binny and Saha, Punyajoy and Yimam, Seid Muhie and Biemann, Chris and Goyal, Pawan and Mukherjee, Animesh},
  booktitle={Proceedings of the AAAI conference on artificial intelligence},
  volume={35},
  pages={14867--14875},
  year={2021},
  doi={10.1609/aaai.v35i17.17745}
}

@inproceedings{hartvigsen-etal-2022-toxigen,
    title = "{T}oxi{G}en: A Large-Scale Machine-Generated Dataset for Adversarial and Implicit Hate Speech Detection",
    author = "Hartvigsen, Thomas  and
      Gabriel, Saadia  and
      Palangi, Hamid  and
      Sap, Maarten  and
      Ray, Dipankar  and
      Kamar, Ece",
    editor = "Muresan, Smaranda  and
      Nakov, Preslav  and
      Villavicencio, Aline",
    booktitle = "Proceedings of the 60th Annual Meeting of the Association for Computational Linguistics (Volume 1: Long Papers)",
    month = may,
    year = "2022",
    address = "Dublin, Ireland",
    publisher = "Association for Computational Linguistics",
    url = "https://aclanthology.org/2022.acl-long.234/",
    doi = "10.18653/v1/2022.acl-long.234",
    pages = "3309--3326"
}

@inproceedings{10.1145/3331184.3331262,
author = {Arango, Aym\'{e} and P\'{e}rez, Jorge and Poblete, Barbara},
title = {Hate Speech Detection is Not as Easy as You May Think: A Closer Look at Model Validation},
year = {2019},
isbn = {9781450361729},
publisher = {Association for Computing Machinery},
address = {New York, NY, USA},
url = {https://doi.org/10.1145/3331184.3331262},
doi = {10.1145/3331184.3331262},
booktitle = {Proceedings of the 42nd International ACM SIGIR Conference on Research and Development in Information Retrieval},
pages = {45–54},
numpages = {10},
location = {Paris, France},
series = {SIGIR'19}
}

@article{chhabra2023literature,
  title={A literature survey on multimodal and multilingual automatic hate speech identification},
  author={Chhabra, Anusha and Vishwakarma, Dinesh Kumar},
  journal={Multimedia Systems},
  volume={29},
  number={3},
  pages={1203--1230},
  year={2023},
  publisher={Springer},
  doi={10.1007/s00530-023-01051-8}
}

@ARTICLE{10662891,
  author={Hashmi, Ehtesham and Yildirim Yayilgan, Sule and Hameed, Ibrahim A. and Mudassar Yamin, Muhammad and Ullah, Mohib and Abomhara, Mohamed},
  journal={IEEE Access}, 
  title={Enhancing Multilingual Hate Speech Detection: From Language-Specific Insights to Cross-Linguistic Integration}, 
  year={2024},
  volume={12},
  number={},
  pages={121507-121537},
  doi={10.1109/ACCESS.2024.3452987}}

@inproceedings{rottger-etal-2022-multilingual,
    title = "Multilingual {H}ate{C}heck: Functional Tests for Multilingual Hate Speech Detection Models",
    author = {R{\"o}ttger, Paul  and
      Seelawi, Haitham  and
      Nozza, Debora  and
      Talat, Zeerak  and
      Vidgen, Bertie},
    editor = "Narang, Kanika  and
      Mostafazadeh Davani, Aida  and
      Mathias, Lambert  and
      Vidgen, Bertie  and
      Talat, Zeerak",
    booktitle = "Proceedings of the Sixth Workshop on Online Abuse and Harms (WOAH)",
    month = jul,
    year = "2022",
    address = "Seattle, Washington (Hybrid)",
    publisher = "Association for Computational Linguistics",
    url = "https://aclanthology.org/2022.woah-1.15/",
    doi = "10.18653/v1/2022.woah-1.15",
    pages = "154--169"
}

@inproceedings{arango-monnar-etal-2022-resources,
    title = "Resources for Multilingual Hate Speech Detection",
    author = "Arango Monnar, Ayme  and
      Perez, Jorge  and
      Poblete, Barbara  and
      Salda{\~n}a, Magdalena  and
      Proust, Valentina",
    editor = "Narang, Kanika  and
      Mostafazadeh Davani, Aida  and
      Mathias, Lambert  and
      Vidgen, Bertie  and
      Talat, Zeerak",
    booktitle = "Proceedings of the Sixth Workshop on Online Abuse and Harms (WOAH)",
    month = jul,
    year = "2022",
    address = "Seattle, Washington (Hybrid)",
    publisher = "Association for Computational Linguistics",
    url = "https://aclanthology.org/2022.woah-1.12/",
    doi = "10.18653/v1/2022.woah-1.12",
    pages = "122--130"
}

@INPROCEEDINGS{9093414,
  author={Gomez, Raul and Gibert, Jaume and Gomez, Lluis and Karatzas, Dimosthenis},
  booktitle={2020 IEEE Winter Conference on Applications of Computer Vision (WACV)}, 
  title={Exploring Hate Speech Detection in Multimodal Publications}, 
  year={2020},
  volume={},
  number={},
  pages={1459-1467},
  doi={10.1109/WACV45572.2020.9093414}}

@inproceedings{bui-etal-2025-multi3hate,
    title = "{M}ulti$^3${H}ate: Multimodal, Multilingual, and Multicultural Hate Speech Detection with Vision{--}Language Models",
    author = "Bui, Minh Duc  and
      Wense, Katharina Von Der  and
      Lauscher, Anne",
    editor = "Chiruzzo, Luis  and
      Ritter, Alan  and
      Wang, Lu",
    booktitle = "Proceedings of the 2025 Conference of the Nations of the Americas Chapter of the Association for Computational Linguistics: Human Language Technologies (Volume 1: Long Papers)",
    month = apr,
    year = "2025",
    address = "Albuquerque, New Mexico",
    publisher = "Association for Computational Linguistics",
    url = "https://aclanthology.org/2025.naacl-long.490/",
    doi = "10.18653/v1/2025.naacl-long.490",
    pages = "9714--9731",
    ISBN = "979-8-89176-189-6"
}

@inproceedings{hee-etal-2024-recent,
    title = "Recent Advances in Online Hate Speech Moderation: Multimodality and the Role of Large Models",
    author = "Hee, Ming Shan  and
      Sharma, Shivam  and
      Cao, Rui  and
      Nandi, Palash  and
      Nakov, Preslav  and
      Chakraborty, Tanmoy  and
      Lee, Roy Ka-Wei",
    editor = "Al-Onaizan, Yaser  and
      Bansal, Mohit  and
      Chen, Yun-Nung",
    booktitle = "Findings of the Association for Computational Linguistics: EMNLP 2024",
    month = nov,
    year = "2024",
    address = "Miami, Florida, USA",
    publisher = "Association for Computational Linguistics",
    url = "https://aclanthology.org/2024.findings-emnlp.254/",
    doi = "10.18653/v1/2024.findings-emnlp.254",
    pages = "4407--4419"
}

@inproceedings{ren-etal-2025-measure,
    title = "How do we measure privacy in text? A survey of text anonymization metrics",
    author = "Ren, Yaxuan  and
      Ramesh, Krithika  and
      Yao, Yaxing  and
      Field, Anjalie",
    editor = "Inui, Kentaro  and
      Sakti, Sakriani  and
      Wang, Haofen  and
      Wong, Derek F.  and
      Bhattacharyya, Pushpak  and
      Banerjee, Biplab  and
      Ekbal, Asif  and
      Chakraborty, Tanmoy  and
      Singh, Dhirendra Pratap",
    booktitle = "Proceedings of the 14th International Joint Conference on Natural Language Processing and the 4th Conference of the Asia-Pacific Chapter of the Association for Computational Linguistics",
    month = dec,
    year = "2025",
    address = "Mumbai, India",
    publisher = "The Asian Federation of Natural Language Processing and The Association for Computational Linguistics",
    url = "https://aclanthology.org/2025.findings-ijcnlp.94/",
    doi = "10.18653/v1/2025.findings-ijcnlp.94",
    pages = "1532--1544",
    ISBN = "979-8-89176-303-6"
}

@INPROCEEDINGS{11247969,
  author={Deußer, Tobias and Sparrenberg, Lorenz and Berger, Armin and Hahnbück, Max and Bauckhage, Christian and Sifa, Rafet},
  booktitle={2025 IEEE 12th International Conference on Data Science and Advanced Analytics (DSAA)}, 
  title={A Survey on Current Trends and Recent Advances in Text Anonymization}, 
  year={2025},
  volume={},
  number={},
  pages={1-9},
  doi={10.1109/DSAA65442.2025.11247969}}

@inproceedings{meisenbacher-etal-2024-comparative,
    title = "A Comparative Analysis of Word-Level Metric Differential Privacy: Benchmarking the Privacy-Utility Trade-off",
    author = "Meisenbacher, Stephen  and
      Nandakumar, Nihildev  and
      Klymenko, Alexandra  and
      Matthes, Florian",
    editor = "Calzolari, Nicoletta  and
      Kan, Min-Yen  and
      Hoste, Veronique  and
      Lenci, Alessandro  and
      Sakti, Sakriani  and
      Xue, Nianwen",
    booktitle = "Proceedings of the 2024 Joint International Conference on Computational Linguistics, Language Resources and Evaluation (LREC-COLING 2024)",
    month = may,
    year = "2024",
    address = "Torino, Italia",
    publisher = "ELRA and ICCL",
    url = "https://aclanthology.org/2024.lrec-main.16/",
    pages = "174--185"
}

@article{10.1145/2382448.2382450,
author = {Brennan, Michael and Afroz, Sadia and Greenstadt, Rachel},
title = {Adversarial stylometry: Circumventing authorship recognition to preserve privacy and anonymity},
year = {2012},
issue_date = {November 2012},
publisher = {Association for Computing Machinery},
address = {New York, NY, USA},
volume = {15},
number = {3},
issn = {1094-9224},
url = {https://doi.org/10.1145/2382448.2382450},
doi = {10.1145/2382448.2382450},
journal = {ACM Trans. Inf. Syst. Secur.},
month = nov,
articleno = {12},
numpages = {22}
}

@inproceedings{baroud-etal-2025-beyond,
    title = "Beyond De-Identification: A Structured Approach for Defining and Detecting Indirect Identifiers in Medical Texts",
    author = {Baroud, Ibrahim  and
      Raithel, Lisa  and
      M{\"o}ller, Sebastian  and
      Roller, Roland},
    editor = "Habernal, Ivan  and
      Ghanavati, Sepideh  and
      Jain, Vijayanta  and
      Igamberdiev, Timour  and
      Wilson, Shomir",
    booktitle = "Proceedings of the Sixth Workshop on Privacy in Natural Language Processing",
    month = apr,
    year = "2025",
    address = "Albuquerque, New Mexico",
    publisher = "Association for Computational Linguistics",
    url = "https://aclanthology.org/2025.privatenlp-main.7/",
    doi = "10.18653/v1/2025.privatenlp-main.7",
    pages = "75--85",
    ISBN = "979-8-89176-246-6"
}

@inproceedings{klymenko-etal-2022-differential,
    title = "Differential Privacy in Natural Language Processing: The Story So Far",
    author = "Klymenko, Oleksandra  and
      Meisenbacher, Stephen  and
      Matthes, Florian",
    editor = "Feyisetan, Oluwaseyi  and
      Ghanavati, Sepideh  and
      Thaine, Patricia  and
      Habernal, Ivan  and
      Mireshghallah, Fatemehsadat",
    booktitle = "Proceedings of the Fourth Workshop on Privacy in Natural Language Processing",
    month = jul,
    year = "2022",
    address = "Seattle, United States",
    publisher = "Association for Computational Linguistics",
    url = "https://aclanthology.org/2022.privatenlp-1.1/",
    doi = "10.18653/v1/2022.privatenlp-1.1",
    pages = "1--11"
}

@inproceedings{yang-etal-2025-robust,
    title = "Robust Utility-Preserving Text Anonymization Based on Large Language Models",
    author = "Yang, Tianyu  and
      Zhu, Xiaodan  and
      Gurevych, Iryna",
    editor = "Che, Wanxiang  and
      Nabende, Joyce  and
      Shutova, Ekaterina  and
      Pilehvar, Mohammad Taher",
    booktitle = "Proceedings of the 63rd Annual Meeting of the Association for Computational Linguistics (Volume 1: Long Papers)",
    month = jul,
    year = "2025",
    address = "Vienna, Austria",
    publisher = "Association for Computational Linguistics",
    url = "https://aclanthology.org/2025.acl-long.1404/",
    doi = "10.18653/v1/2025.acl-long.1404",
    pages = "28922--28941",
    ISBN = "979-8-89176-251-0"
}

@inproceedings{frikha-etal-2025-incognitext,
    title = "{I}ncogni{T}ext: Privacy-enhancing Conditional Text Anonymization via {LLM}-based Private Attribute Randomization",
    author = "Frikha, Ahmed  and
      Walha, Nassim  and
      Nakka, Krishna Kanth  and
      Mendes, Ricardo  and
      Jiang, Xue  and
      Zhou, Xuebing",
    editor = "Inui, Kentaro  and
      Sakti, Sakriani  and
      Wang, Haofen  and
      Wong, Derek F.  and
      Bhattacharyya, Pushpak  and
      Banerjee, Biplab  and
      Ekbal, Asif  and
      Chakraborty, Tanmoy  and
      Singh, Dhirendra Pratap",
    booktitle = "Proceedings of the 14th International Joint Conference on Natural Language Processing and the 4th Conference of the Asia-Pacific Chapter of the Association for Computational Linguistics",
    month = dec,
    year = "2025",
    address = "Mumbai, India",
    publisher = "The Asian Federation of Natural Language Processing and The Association for Computational Linguistics",
    url = "https://aclanthology.org/2025.ijcnlp-long.134/",
    doi = "10.18653/v1/2025.ijcnlp-long.134",
    pages = "2490--2501",
    ISBN = "979-8-89176-298-5"
}

@inproceedings{carvalho2023tem,
  title={TEM: High utility metric differential privacy on text},
  author={Carvalho, Ricardo Silva and Vasiloudis, Theodore and Feyisetan, Oluwaseyi and Wang, Ke},
  booktitle={Proceedings of the 2023 SIAM International Conference on Data Mining (SDM)},
  pages={883--890},
  year={2023},
  organization={SIAM},
  doi={10.1137/1.9781611977653.ch99}
}

@inproceedings{igamberdiev-habernal-2023-dp,
    title = "{DP}-{BART} for Privatized Text Rewriting under Local Differential Privacy",
    author = "Igamberdiev, Timour  and
      Habernal, Ivan",
    editor = "Rogers, Anna  and
      Boyd-Graber, Jordan  and
      Okazaki, Naoaki",
    booktitle = "Findings of the Association for Computational Linguistics: ACL 2023",
    month = jul,
    year = "2023",
    address = "Toronto, Canada",
    publisher = "Association for Computational Linguistics",
    url = "https://aclanthology.org/2023.findings-acl.874/",
    doi = "10.18653/v1/2023.findings-acl.874",
    pages = "13914--13934"
}

@inproceedings{bao-carpuat-2024-keep,
    title = "{K}eep it {P}rivate: Unsupervised Privatization of Online Text",
    author = "Bao, Calvin  and
      Carpuat, Marine",
    editor = "Duh, Kevin  and
      Gomez, Helena  and
      Bethard, Steven",
    booktitle = "Proceedings of the 2024 Conference of the North American Chapter of the Association for Computational Linguistics: Human Language Technologies (Volume 1: Long Papers)",
    month = jun,
    year = "2024",
    address = "Mexico City, Mexico",
    publisher = "Association for Computational Linguistics",
    url = "https://aclanthology.org/2024.naacl-long.480/",
    doi = "10.18653/v1/2024.naacl-long.480",
    pages = "8678--8693"
}

@inproceedings{10.1145/3531146.3534642,
author = {Brown, Hannah and Lee, Katherine and Mireshghallah, Fatemehsadat and Shokri, Reza and Tram\`{e}r, Florian},
title = {What Does it Mean for a Language Model to Preserve Privacy?},
year = {2022},
isbn = {9781450393522},
publisher = {Association for Computing Machinery},
address = {New York, NY, USA},
url = {https://doi.org/10.1145/3531146.3534642},
doi = {10.1145/3531146.3534642},
booktitle = {Proceedings of the 2022 ACM Conference on Fairness, Accountability, and Transparency},
pages = {2280–2292},
numpages = {13},
location = {Seoul, Republic of Korea},
series = {FAccT '22}
}

@inproceedings{loiseau-etal-2025-tau,
    title = "Tau-Eval: A Unified Evaluation Framework for Useful and Private Text Anonymization",
    author = "Loiseau, Gabriel  and
      Sileo, Damien  and
      Riquet, Damien  and
      Meyer, Maxime  and
      Tommasi, Marc",
    editor = {Habernal, Ivan  and
      Schulam, Peter  and
      Tiedemann, J{\"o}rg},
    booktitle = "Proceedings of the 2025 Conference on Empirical Methods in Natural Language Processing: System Demonstrations",
    month = nov,
    year = "2025",
    address = "Suzhou, China",
    publisher = "Association for Computational Linguistics",
    url = "https://aclanthology.org/2025.emnlp-demos.16/",
    doi = "10.18653/v1/2025.emnlp-demos.16",
    pages = "216--227",
    ISBN = "979-8-89176-334-0"
}

@article{pilan-etal-2022-text,
    title = "The Text Anonymization Benchmark ({TAB}): A Dedicated Corpus and Evaluation Framework for Text Anonymization",
    author = "Pil{\'a}n, Ildik{\'o}  and
      Lison, Pierre  and
      {\O}vrelid, Lilja  and
      Papadopoulou, Anthi  and
      S{\'a}nchez, David  and
      Batet, Montserrat",
    journal = "Computational Linguistics",
    volume = "48",
    number = "4",
    month = dec,
    year = "2022",
    address = "Cambridge, MA",
    publisher = "MIT Press",
    url = "https://aclanthology.org/2022.cl-4.19/",
    doi = "10.1162/coli_a_00458",
    pages = "1053--1101"
}

@inproceedings{huang-etal-2025-nap2,
    title = "{NAP}2: A Benchmark for Naturalness and Privacy-Preserving Text Rewriting by Learning from Human",
    author = "Huang, Shuo  and
      Maclean, William  and
      Kang, Xiaoxi  and
      Xu, Qiongkai  and
      Li, Zhuang  and
      Yuan, Xingliang  and
      Haffari, Gholamreza  and
      Qu, Lizhen",
    editor = "Christodoulopoulos, Christos  and
      Chakraborty, Tanmoy  and
      Rose, Carolyn  and
      Peng, Violet",
    booktitle = "Findings of the Association for Computational Linguistics: EMNLP 2025",
    month = nov,
    year = "2025",
    address = "Suzhou, China",
    publisher = "Association for Computational Linguistics",
    url = "https://aclanthology.org/2025.findings-emnlp.476/",
    doi = "10.18653/v1/2025.findings-emnlp.476",
    pages = "8954--8970",
    ISBN = "979-8-89176-335-7"
}

@inproceedings{qian-etal-2019-benchmark,
    title = "A Benchmark Dataset for Learning to Intervene in Online Hate Speech",
    author = "Qian, Jing  and
      Bethke, Anna  and
      Liu, Yinyin  and
      Belding, Elizabeth  and
      Wang, William Yang",
    editor = "Inui, Kentaro  and
      Jiang, Jing  and
      Ng, Vincent  and
      Wan, Xiaojun",
    booktitle = "Proceedings of the 2019 Conference on Empirical Methods in Natural Language Processing and the 9th International Joint Conference on Natural Language Processing (EMNLP-IJCNLP)",
    month = nov,
    year = "2019",
    address = "Hong Kong, China",
    publisher = "Association for Computational Linguistics",
    url = "https://aclanthology.org/D19-1482/",
    doi = "10.18653/v1/D19-1482",
    pages = "4755--4764"
}

@inproceedings{waseem-hovy-2016-hateful,
    title = "Hateful Symbols or Hateful People? Predictive Features for Hate Speech Detection on {T}witter",
    author = "Waseem, Zeerak  and
      Hovy, Dirk",
    editor = "Andreas, Jacob  and
      Choi, Eunsol  and
      Lazaridou, Angeliki",
    booktitle = "Proceedings of the {NAACL} Student Research Workshop",
    month = jun,
    year = "2016",
    address = "San Diego, California",
    publisher = "Association for Computational Linguistics",
    url = "https://aclanthology.org/N16-2013/",
    doi = "10.18653/v1/N16-2013",
    pages = "88--93"
}

@inproceedings{devlin-etal-2019-bert,
    title = "{BERT}: Pre-training of Deep Bidirectional Transformers for Language Understanding",
    author = "Devlin, Jacob  and
      Chang, Ming-Wei  and
      Lee, Kenton  and
      Toutanova, Kristina",
    editor = "Burstein, Jill  and
      Doran, Christy  and
      Solorio, Thamar",
    booktitle = "Proceedings of the 2019 Conference of the North {A}merican Chapter of the Association for Computational Linguistics: Human Language Technologies, Volume 1 (Long and Short Papers)",
    month = jun,
    year = "2019",
    address = "Minneapolis, Minnesota",
    publisher = "Association for Computational Linguistics",
    url = "https://aclanthology.org/N19-1423/",
    doi = "10.18653/v1/N19-1423",
    pages = "4171--4186"
}

@inproceedings{mattern-etal-2022-limits,
    title = "The Limits of Word Level Differential Privacy",
    author = "Mattern, Justus  and
      Weggenmann, Benjamin  and
      Kerschbaum, Florian",
    editor = "Carpuat, Marine  and
      de Marneffe, Marie-Catherine  and
      Meza Ruiz, Ivan Vladimir",
    booktitle = "Findings of the Association for Computational Linguistics: NAACL 2022",
    month = jul,
    year = "2022",
    address = "Seattle, United States",
    publisher = "Association for Computational Linguistics",
    url = "https://aclanthology.org/2022.findings-naacl.65/",
    doi = "10.18653/v1/2022.findings-naacl.65",
    pages = "867--881"
}

@inproceedings{zaratiana-etal-2024-gliner,
    title = "{GL}i{NER}: Generalist Model for Named Entity Recognition using Bidirectional Transformer",
    author = "Zaratiana, Urchade  and
      Tomeh, Nadi  and
      Holat, Pierre  and
      Charnois, Thierry",
    editor = "Duh, Kevin  and
      Gomez, Helena  and
      Bethard, Steven",
    booktitle = "Proceedings of the 2024 Conference of the North American Chapter of the Association for Computational Linguistics: Human Language Technologies (Volume 1: Long Papers)",
    month = jun,
    year = "2024",
    address = "Mexico City, Mexico",
    publisher = "Association for Computational Linguistics",
    url = "https://aclanthology.org/2024.naacl-long.300/",
    doi = "10.18653/v1/2024.naacl-long.300",
    pages = "5364--5376"
}

@inproceedings{yue-etal-2021-differential,
    title = "Differential Privacy for Text Analytics via Natural Text Sanitization",
    author = "Yue, Xiang  and
      Du, Minxin  and
      Wang, Tianhao  and
      Li, Yaliang  and
      Sun, Huan  and
      Chow, Sherman S. M.",
    editor = "Zong, Chengqing  and
      Xia, Fei  and
      Li, Wenjie  and
      Navigli, Roberto",
    booktitle = "Findings of the Association for Computational Linguistics: ACL-IJCNLP 2021",
    month = aug,
    year = "2021",
    address = "Online",
    publisher = "Association for Computational Linguistics",
    url = "https://aclanthology.org/2021.findings-acl.337/",
    doi = "10.18653/v1/2021.findings-acl.337",
    pages = "3853--3866"
}

@inproceedings{meisenbacher-etal-2024-dp,
    title = "{DP}-{MLM}: Differentially Private Text Rewriting Using Masked Language Models",
    author = "Meisenbacher, Stephen  and
      Chevli, Maulik  and
      Vladika, Juraj  and
      Matthes, Florian",
    editor = "Ku, Lun-Wei  and
      Martins, Andre  and
      Srikumar, Vivek",
    booktitle = "Findings of the Association for Computational Linguistics: ACL 2024",
    month = aug,
    year = "2024",
    address = "Bangkok, Thailand",
    publisher = "Association for Computational Linguistics",
    url = "https://aclanthology.org/2024.findings-acl.554/",
    doi = "10.18653/v1/2024.findings-acl.554",
    pages = "9314--9328"
}

@article{radford2019language,
  title={Language Models are Unsupervised Multitask Learners},
  author={Radford, Alec and Wu, Jeff and Child, Rewon and Luan, David and Amodei, Dario and Sutskever, Ilya},
  year={2019},
  journal={OpenAI},
  url={https://cdn.openai.com/better-language-models/language_models_are_unsupervised_multitask_learners.pdf}
}

@ARTICLE{9455353,
  author={Mullah, Nanlir Sallau and Zainon, Wan Mohd Nazmee Wan},
  journal={IEEE Access}, 
  title={Advances in Machine Learning Algorithms for Hate Speech Detection in Social Media: A Review}, 
  year={2021},
  volume={9},
  number={},
  pages={88364-88376},
  doi={10.1109/ACCESS.2021.3089515}}

@article{tucudean2024natural,
  title={Natural language processing with transformers: a review},
  author={Tucudean, Georgiana and Bucos, Marian and Dragulescu, Bogdan and Caleanu, Catalin Daniel},
  journal={PeerJ Computer Science},
  volume={10},
  pages={e2222},
  year={2024},
  publisher={PeerJ Inc.},
  doi={10.7717/peerj-cs.2222}
}

@inproceedings{loiseau-etal-2025-tarot,
    title = "{TAROT}: Task-Oriented Authorship Obfuscation Using Policy Optimization Methods",
    author = "Loiseau, Gabriel  and
      Sileo, Damien  and
      Riquet, Damien  and
      Meyer, Maxime  and
      Tommasi, Marc",
    editor = "Habernal, Ivan  and
      Ghanavati, Sepideh  and
      Jain, Vijayanta  and
      Igamberdiev, Timour  and
      Wilson, Shomir",
    booktitle = "Proceedings of the Sixth Workshop on Privacy in Natural Language Processing",
    month = apr,
    year = "2025",
    address = "Albuquerque, New Mexico",
    publisher = "Association for Computational Linguistics",
    url = "https://aclanthology.org/2025.privatenlp-main.2/",
    doi = "10.18653/v1/2025.privatenlp-main.2",
    pages = "14--31",
    ISBN = "979-8-89176-246-6"
}

@article{potthast2016author,
  title={Author Obfuscation: Attacking the State of the Art in Authorship Verification.},
  author={Potthast, Martin and Hagen, Matthias and Stein, Benno},
  journal={CLEF (Working Notes)},
  pages={716--749},
  year={2016},
  url={https://ceur-ws.org/Vol-1609/16090716.pdf}
}

@inproceedings{bevendorff-etal-2019-heuristic,
    title = "Heuristic Authorship Obfuscation",
    author = "Bevendorff, Janek  and
      Potthast, Martin  and
      Hagen, Matthias  and
      Stein, Benno",
    editor = "Korhonen, Anna  and
      Traum, David  and
      M{\`a}rquez, Llu{\'i}s",
    booktitle = "Proceedings of the 57th Annual Meeting of the Association for Computational Linguistics",
    month = jul,
    year = "2019",
    address = "Florence, Italy",
    publisher = "Association for Computational Linguistics",
    url = "https://aclanthology.org/P19-1104/",
    doi = "10.18653/v1/P19-1104",
    pages = "1098--1108"
}

@inproceedings{fisher-etal-2024-styleremix,
    title = "{S}tyle{R}emix: Interpretable Authorship Obfuscation via Distillation and Perturbation of Style Elements",
    author = "Fisher, Jillian  and
      Hallinan, Skyler  and
      Lu, Ximing  and
      Gordon, Mitchell L  and
      Harchaoui, Zaid  and
      Choi, Yejin",
    editor = "Al-Onaizan, Yaser  and
      Bansal, Mohit  and
      Chen, Yun-Nung",
    booktitle = "Proceedings of the 2024 Conference on Empirical Methods in Natural Language Processing",
    month = nov,
    year = "2024",
    address = "Miami, Florida, USA",
    publisher = "Association for Computational Linguistics",
    url = "https://aclanthology.org/2024.emnlp-main.241/",
    doi = "10.18653/v1/2024.emnlp-main.241",
    pages = "4172--4206"
}

@article{10.1145/3715073.3715076,
author = {Huang, Baixiang and Chen, Canyu and Shu, Kai},
title = {Authorship Attribution in the Era of LLMs: Problems, Methodologies, and Challenges},
year = {2025},
issue_date = {December 2024},
publisher = {Association for Computing Machinery},
address = {New York, NY, USA},
volume = {26},
number = {2},
issn = {1931-0145},
url = {https://doi.org/10.1145/3715073.3715076},
doi = {10.1145/3715073.3715076},
journal = {SIGKDD Explor. Newsl.},
month = jan,
pages = {21–43},
numpages = {23}
}

@inproceedings{shokri-etal-2025-personalized,
    title = "Personalized Author Obfuscation with Large Language Models",
    author = "Shokri, Mohammad  and
      Levitan, Sarah Ita  and
      Levitan, Rivka",
    editor = "Angelova, Galia  and
      Kunilovskaya, Maria  and
      Escribe, Marie  and
      Mitkov, Ruslan",
    booktitle = "Proceedings of the 15th International Conference on Recent Advances in Natural Language Processing - Natural Language Processing in the Generative AI Era",
    month = sep,
    year = "2025",
    address = "Varna, Bulgaria",
    publisher = "INCOMA Ltd., Shoumen, Bulgaria",
    url = "https://aclanthology.org/2025.ranlp-1.133/",
    pages = "1153--1162"
}

@inproceedings{stamatatos-2017-authorship,
    title = "Authorship Attribution Using Text Distortion",
    author = "Stamatatos, Efstathios",
    editor = "Lapata, Mirella  and
      Blunsom, Phil  and
      Koller, Alexander",
    booktitle = "Proceedings of the 15th Conference of the {E}uropean Chapter of the Association for Computational Linguistics: Volume 1, Long Papers",
    month = apr,
    year = "2017",
    address = "Valencia, Spain",
    publisher = "Association for Computational Linguistics",
    url = "https://aclanthology.org/E17-1107/",
    pages = "1138--1149"
}

@inproceedings{tan-etal-2025-open,
    title = "Open-World Authorship Attribution",
    author = "Tan, Xinhao  and
      Liu, Songhua  and
      Cong, Xia  and
      Li, Kunjun  and
      Wang, Xinchao",
    editor = "Che, Wanxiang  and
      Nabende, Joyce  and
      Shutova, Ekaterina  and
      Pilehvar, Mohammad Taher",
    booktitle = "Findings of the Association for Computational Linguistics: ACL 2025",
    month = jul,
    year = "2025",
    address = "Vienna, Austria",
    publisher = "Association for Computational Linguistics",
    url = "https://aclanthology.org/2025.findings-acl.913/",
    doi = "10.18653/v1/2025.findings-acl.913",
    pages = "17744--17758",
    ISBN = "979-8-89176-256-5"
}

@inproceedings{bevendorff-etal-2025-two,
    title = "The Two Paradigms of {LLM} Detection: Authorship Attribution vs. Authorship Verification",
    author = "Bevendorff, Janek  and
      Wiegmann, Matti  and
      Richter, Emmelie  and
      Potthast, Martin  and
      Stein, Benno",
    editor = "Che, Wanxiang  and
      Nabende, Joyce  and
      Shutova, Ekaterina  and
      Pilehvar, Mohammad Taher",
    booktitle = "Findings of the Association for Computational Linguistics: ACL 2025",
    month = jul,
    year = "2025",
    address = "Vienna, Austria",
    publisher = "Association for Computational Linguistics",
    url = "https://aclanthology.org/2025.findings-acl.194/",
    doi = "10.18653/v1/2025.findings-acl.194",
    pages = "3762--3787",
    ISBN = "979-8-89176-256-5"
}

@inproceedings{hung-etal-2023-wrote,
    title = "Who Wrote it and Why? Prompting Large-Language Models for Authorship Verification",
    author = "Hung, Chia-Yu  and
      Hu, Zhiqiang  and
      Hu, Yujia  and
      Lee, Roy",
    editor = "Bouamor, Houda  and
      Pino, Juan  and
      Bali, Kalika",
    booktitle = "Findings of the Association for Computational Linguistics: EMNLP 2023",
    month = dec,
    year = "2023",
    address = "Singapore",
    publisher = "Association for Computational Linguistics",
    url = "https://aclanthology.org/2023.findings-emnlp.937/",
    doi = "10.18653/v1/2023.findings-emnlp.937",
    pages = "14078--14084"
}

@inproceedings{mishra-etal-2018-author,
    title = "Author Profiling for Abuse Detection",
    author = "Mishra, Pushkar  and
      Del Tredici, Marco  and
      Yannakoudakis, Helen  and
      Shutova, Ekaterina",
    editor = "Bender, Emily M.  and
      Derczynski, Leon  and
      Isabelle, Pierre",
    booktitle = "Proceedings of the 27th International Conference on Computational Linguistics",
    month = aug,
    year = "2018",
    address = "Santa Fe, New Mexico, USA",
    publisher = "Association for Computational Linguistics",
    url = "https://aclanthology.org/C18-1093/",
    pages = "1088--1098"
}

\end{document}